\documentclass[10pt,twocolumn,letterpaper]{article}

\usepackage{cvpr}              % To produce the CAMERA-READY version
\usepackage{graphicx}
\usepackage{amsmath}
\usepackage{amssymb}
\usepackage{booktabs}
\usepackage{multicol}
\usepackage{color}
\usepackage{multirow}
\usepackage{makecell}
\usepackage{soul}
\usepackage{comment}

\usepackage[pagebackref,breaklinks,colorlinks]{hyperref}

\usepackage[capitalize]{cleveref}
\crefname{section}{Sec.}{Secs.}
\Crefname{section}{Section}{Sections}
\Crefname{table}{Table}{Tables}
\crefname{table}{Tab.}{Tabs.}

\def\cvprPaperID{20} % *** Enter the CVPR Paper ID here
\def\confName{CVPR}
\def\confYear{2023}

\begin{document}
	% \colorlet{mygreen}{green!60!gray}
	
	\sloppy
	\definecolor{black}{rgb}{0.0, 0.0, 0.0}
	\definecolor{white}{rgb}{1.0, 1.0, 1.0}
	\definecolor{yellow}{rgb}{1.0, 1.0, 0.8}
	\definecolor{blue}{rgb}{0.0, 0.2, 0.5}
	\definecolor{green}{rgb}{0.6, 0.8, 0.8}
	\definecolor{dark_green}{RGB} {0, 140, 0}
	\definecolor{gold}{rgb}{0.6, 0.4, 0.1}
	\definecolor{grey}{RGB}{0,0,0}
	\definecolor{Gray}{gray}{0.8}
	\definecolor{MediumGray}{gray}{0.9}
	\definecolor{LightGray}{gray}{0.98}
	%\definecolor{LightCyan}{rgb}{0.88,1,1}
	\definecolor{purple}{RGB}{128,0,128}
	\definecolor{sl_blue}{RGB}{47, 60, 105}
	\definecolor{orange}{RGB}{255,165,0}
	\definecolor{Gray}{gray}{0.85}
	
	\newif\ifcomments
	
	% Uncomment one of these two lines
	\commentstrue  % uncomment to display comments in text
	 % \commentsfalse % uncomment to hide comments in text
	
	\ifcomments
	\newcommand{\JUN}[1]{\textcolor{orange}{\textbf{\small [}\colorbox{yellow}{\textbf{Jun:}}{\small #1}\textbf{\small ]}}}
	\newcommand{\MB}[1]{\textcolor{dark_green}{\textbf{\small [}\colorbox{yellow}{\textbf{Mauro:}}{\small #1}\textbf{\small ]}}}
	\newcommand{\OMRAN}[1]{\textcolor{blue}{\textbf{\small [}\colorbox{yellow}{\textbf{Omran:}}{\small #1}\textbf{\small ]}}}
	\newcommand{\BTcomm}[1]{\textcolor{purple}{\textbf{\small [}\colorbox{yellow}{\textbf{Benedetta:}}{\small #1}\textbf{\small ]}}}
	\newcommand{\TODO}[1]{\textcolor{red}{{TODO: #1}}}
	\newcommand{\CH}[1]{\textcolor{brown}{{#1}}}
	\newcommand{\stm}[1]{\textcolor{red}{{\st{#1}}}}
	\else
	\newcommand{\JUN}[1]{}
	\newcommand{\MB}[1]{}
	\newcommand{\OMRAN}[1]{}
	\newcommand{\BTcomm}[1]{}
	\newcommand{\BT}[1]{}
	\newcommand{\TODO}[1]{}
	\newcommand{\CH}[1]{}
	\newcommand{\stm}[1]{}
	\fi
	
	\title{Open Set Classification of GAN-based Image Manipulations \\via a ViT-based Hybrid Architecture}

% \title{Open Set Classification of Synthetic Manipulations via a\\ ViT-based Hybrid Architecture}
 
	\author{Jun Wang \quad Omran Alamayreh \quad Benedetta Tondi \quad Mauro Barni\\
		Department of Information Engineering and Mathematics, University of Siena\\
		{\tt\small j.wang@student.unisi.it, omran@diism.unisi.it} %lydia.abady@unisi.it}\\
		{\tt\small benedetta.tondi@unisi.it, barni@dii.unisi.it}\\
	}
	\maketitle
	
	%%%%%%%%% ABSTRACT
	\begin{abstract}
		Classification of AI-manipulated content is receiving great attention, for distinguishing different types of manipulations. Most of the methods developed so far fail in the open-set scenario, that is when the algorithm used for the manipulation is not represented by the training set. 
        In this paper, we focus on the classification of synthetic face generation and manipulation in open-set scenarios, and propose a method for classification with a rejection option. The proposed method combines the use of Vision Transformers (ViT) with a hybrid approach for simultaneous classification and localization. Feature map correlation is exploited by the ViT module, while a localization branch is employed as an attention mechanism to force the model to learn per-class discriminative features associated with the forgery when the manipulation is performed locally in the image. Rejection is performed by considering several strategies and analyzing the model output layers.
		%Extensive experiments on 19 different facial attributes show impressive improvements using the Vit module.
		The effectiveness of the proposed method is assessed for the task of classification of facial attribute editing and GAN attribution.
  
		% \textbf{key words}: face manipulation, GAN attribution, vision transformer, open set recognition, deepfake detection
	\end{abstract}
	
	%%%%%% ----------------------------------------------------------

% \MB{I do not like the tile much, here I propose a number of alternatives}

% {Open Set Classification of GAN-based Image Manipulations via a ViT-based Hybrid Architecture\\}

% {Open Set Classification of facial-attributes Manipulations via a ViT-based Hybrid Architecture\\}

% {A ViT-based Hybrid Architecture for Open Set Classification of Facial-attributes Manipulations\\}

%  A ViT-based Hybrid Architecture for Open Set Classification of GAN-based Image Manipulations\\
 
	\section{Introduction}
	\label{sec:intro}
        Synthetic manipulation of face images has become ubiquitous and is being increasingly used in a wide variety of applications \cite{gui2021review}, thus posing a serious threat to public trust. Many detectors have been proposed to classify images forged by generative models as fakes/synthetic \cite{guarnera2022face}. There are cases where just knowing that the image is fake is not enough and more information is required on the synthetic manipulation undergone by the image.
         This is the case, for instance, when the synthetic manipulation\footnote{In the following, we generically refer to the case of local GAN manipulation and generation from scratch with the term synthetic manipulation.} consists of local attribute editing, rather than in the generation of a synthetic image from from scratch \cite{shen2020interpreting,patashnik2021styleclip}, in which case it is preferable to also provide evidence to support the judgment that the image is fake, rather than simply saying that the image has been manipulated.
         % This is the case, for instance, when the synthetic manipulation\footnote{In the following, we generically refer to the case of local GAN manipulation and generation from scratch with the term synthetic manipulation.} consists in local attribute editing, rather than generation from scratch \cite{shen2020interpreting,patashnik2021styleclip}, in which case it is preferable to
         % also, provide evidence to support the judgment that the image is
         % fake, rather than simply saying that the image has been manipulated.%
         Several methods addressed this problem via binary detectors, judging the image as real or fake, which also have the ability to localize the manipulation, e.g. outputting binary localization masks or attention maps \cite{huang2022fakelocator,zhao2021multi,mazaheri2022detection}, or via multi-class classifiers \cite{wang2022icassp},
         % \BTcomm{Add ref to our ICASSP paper} \BTcomm{No need. Add citation to ICASSP23 conference adding the specification 'to be presented'} \JUN{Done, but I just realized this will give our information to the reviewers. Can we just cite it as ICASSP2023 paper?} \BTcomm{ok. Even if perhaps it does not change much. You could add the arXiv reference now (if you do not have yet a arxiv reference number do not put it) and remember to change on the ICASSP 23, to be presented, in the camera ready.}\JUN{Ok} \BTcomm{Then, please do that and add the other in the camera ready.}, 
         that classify the type of facial attribute editing performed by generators.
         In yet some other cases there is interest in knowing the specific type of architecture used to generate the manipulation (synthetic image attribution), %hence a multi-class classification problem is solved.    	
         Methods have been proposed that perform attribution via multi-class classifiers by relying on artifacts or signatures (fingerprints) left by the models in the generated images \cite{marra2019gans,yang2022deepfake}.
	
        \begin{figure}
            \centering
            \includegraphics[width=\columnwidth]{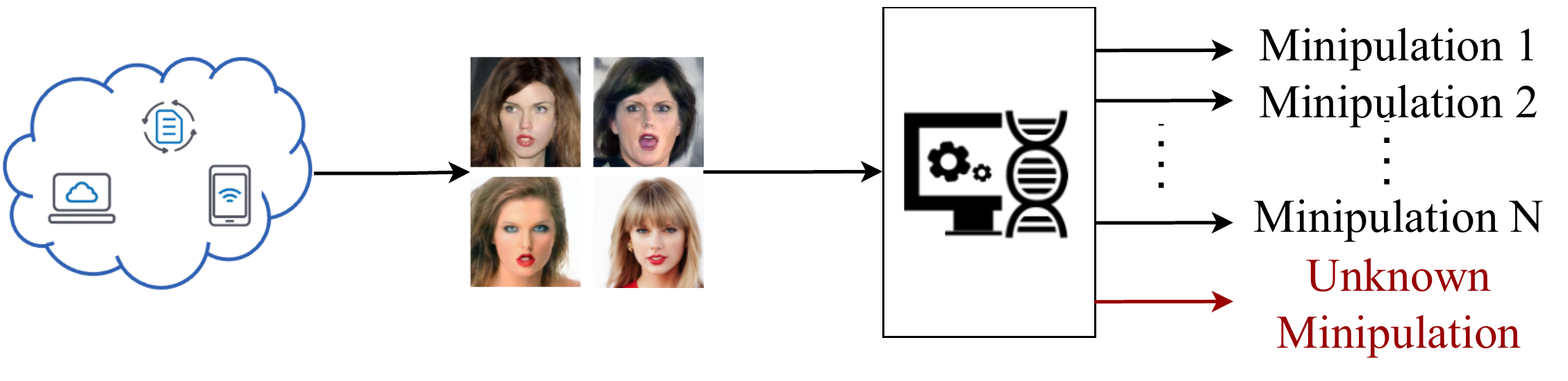}
            \caption{Open set  scenario of synthetic manipulation classification (classification with rejection option)   considered in this paper.}
            \label{fig:scenario}
        \end{figure}
	
        A common drawback with all the above binary and multi-class classification approaches is that their application is limited to closed-set scenarios. For instance, Generative Adversarial Network (GAN) attribution methods can correctly attribute the image only if it comes from a GAN architecture among those seen during training, and they are incapable of identifying or revealing unseen GAN types.
        % (to the best of our knowledge, the only exception is ??? where an algorithm to discover new GANs/architectures from an unlabeled set and clusterise them  is proposed).
        This seriously limits the applicability of these methods in real-world settings, where the images seen during operation time may be edited in different ways or generated by architectures not seen during training, with the consequence that the predictions made by methods are not trustable.
	
        In this paper, we address the problem of classification of synthetic facial manipulation in open set scenarios, proposing a method for classification with rejection option. To build the closed-set classifier, the method combines a Vision Transformer (ViT) with a hybrid approach for simultaneous classification and localization, inspired to \cite{wang2022architecture, wang2022icassp}.
        %\BTcomm{I think that perhaps what we do here is more close to the approach of ClimateGAN since we do not have the spatial patch-based approach that we have in ICASSP....Cite that one in case}\JUN{Done}.
	%, for closed set classification.
        Then, a dedicated module performs rejection/acceptance of the sample by analyzing the model output layers. Rejection is performed by considering the maximum logit score (MLS), maximum softmax probability (MSP), and the OpenMax approach.
        More specifically, in our architecture, the input sequence to the ViT is formed from feature maps extracted from a CNN (\cite{dosovitskiy2020image,wang2022m2tr}), and then the ViT module is used to exploit feature maps correlation, via the self-attention mechanism. In the general case of local manipulation, the same features shared by Vit-based classification heads are also utilized by a localization branch via a fully convolution network. The goal of the localization branch is to force the network to focus on the most significant parts of the image (attention mechanism) \cite{wang2022architecture}.%, that has been shown to have a beneficial effect on the classification accuracy and generalization capability \BTcomm{cite climateGAN}.
        The overall architecture, which includes the feature extraction network, localization branch and ViT module, is then trained in an end-to-end fashion.
        Our method follows some recent works in machine learning, showing that ViT allows achieving improved performance for out-of-distribution detection \cite{fort2021exploring} and open set recognition \cite{al2022open}, compared to standard CNN architectures.
	
	%Main features:
	%Classification with localization branch (hybrid): the requirement to localize helps the classification [*]
	%Exploiting feature maps correlation by ViT module (inspired by [**],[***])
	%via the self-attention mechanism.
        Experiments are carried out considering the classification of facial attribute editing and GAN attribution. For the classification of synthetic facial attributes, we considered 19 editing types, with  manipulations performed by InterfaceGAN \cite{shen2020interpreting} and StyleCLIP \cite{patashnik2021styleclip}. 
        %and different combinations for in-set and  out-of-set manipulations.
         %
        %	Experiments confirm that the proposed architecture and in particular the use of ViT, is beneficial and allows to significantly improve the open set performance for a similar classification accuracy in closed-set, outperforming other state-of-the art methods in the literature for open set recognition (OSR).
	%We also applied the proposed architecture to address the
        For GAN attribution, the performance is assessed considering facial images generated by several modern generative models, namely, LGSM \cite{vahdat2021score}, StyleGAN2 \cite{karras2020analyzing}, StyleGAN3 \cite{karras2021alias}, Taming transformer \cite{esser2021taming} and Latent Diffusion \cite{rombach2022high}.
        %Very good results are achieved for most of  in-set and  out-of-set combinations.
        For both tasks, experiments were performed considering different combinations of in-set and out-of-set manipulations.
        The results confirm that the proposed architecture, and in particular the use of ViT, is beneficial and allows to significantly improve open set performance without impairing the accuracy on closed-set samples, outperforming state-of-the-art methods in the literature for open set recognition (OSR).
	
        %\TODO{Adjust at the end the outline of the paper. We can also remove it} \JUN{Good}
        The rest of the paper is organized as follows. Section \ref{sec:related} introduces the related work on the generation and detection/classification of synthetic faces, and the most relevant methods for open-set classification in machine learning. The proposed architecture is presented in Section \ref{sec:method}. Section \ref{sec:exper} describes the experimental methodology and setting. The results and the comparisons with the state-of-the-art are finally reported and discussed in Section \ref{sec:results}. Finally, we draw conclusions in section \ref{sec:conclusion}.
	
	%%%%%% ----------------------------------------------------------

	\section{Related work}
	\label{sec:related}
	
	\subsection{AI-synthesized faces and their detection}
	Artificial Intelligence (AI)-synthesized faces can be either fully synthetic when the faces are generated from scratch using generative models, or locally manipulated, e.g. when a single facial attribute or multiple attributes are modified by the model while the other attributes remain unchanged.
	
	A wide variety of generative models, notably GANs \cite{karras2020analyzing,karras2021alias} and diffusion models \cite{vahdat2021score,rombach2022high}, are nowadays able to generate high-resolution images from scratch with an unprecedented level of realism.
	Inspired by the superior performance of the StyleGAN series \cite{karras2020analyzing} in synthesizing high-resolution and high-quality images, StyleGAN architectures have been adopted for image editing, achieving high-quality edited images. Among them, we mention InterFaceGAN \cite{shen2020interpreting} and StyleCLIP \cite{radford2021learning}. InterFaceGAN proposes a framework to interpret the disentangled face representation learned by the StyleGAN model and studies the properties of the facial semantics encoded in the latent space, showing that it is possible to edit the semantic attribute through linear subspace projection.
	StyleCLIP is a text-based interface for StyleGAN-based image manipulation. StyleCLIP mainly uses the Contrastive Language-Image Pre-training (CLIP) model to edit the latent code through the user input language description, so as to achieve the purpose of editing the image.
	With regard to the defences, several methods have been developed in the last years to discriminate between fake and real \cite{wang2022m2tr,wang2022eyes,wang2022gan}. Attempts have also been made to attribute the GAN model and/or the type of architecture generating the image. %\cite{yu2019attributing,yang2022deepfake,jia2022model} \BTcomm{cite only one (already cited elsewhere - we need to reduce further the number of references. 2 pages for a conference is really too much!! I already did lot of effort. We need some further cuts) or a survey}.
	%
	%\BTcomm{@Jun: I am drafting the part below. Expand it  (do changing in case)}\JUN{Done}
	In many cases, model-level attribution is performed by relying on the estimation of a fingerprint characterizing the GAN model, e.g.  \cite{yu2019attributing,marra2019gans,zhang2019detecting}. Other works address the task of architecture attribution, attributing the fake image to the source architecture instead of the specific model, see for instance \cite{yang2022deepfake}, where a multi-class classifier is proposed to discriminate among different architectures in a closed-set setting, even when the images are generated by using different initialization, loss and dataset, hence possessing distinct model-level fingerprints. Recently, a method for the classification of the synthetic face editing performed by the GAN has also been proposed in \cite{wang2022icassp}. The method relies on a patch-driven hybrid classification network with localization supervision, that classifies the editing among a pool of possible manipulations (closed-set setting), with good robustness against post-processing. As a drawback of this approach, the pre-training of the patch-based models is time-consuming.

	\subsection{Open Set Recognition}
        \label{sec.related}
	Open set recognition (OSR), first formalized in \cite{Scheirer2013} for classical machine learning, addresses the problem of determining whether an input belongs to one of the classes used to train a network. Such a problem has received increasing attention, especially in the last years \cite{geng2020recent}. A method to address OSR with deep neural network-based approaches, named OpenMax, was presented in \cite{bendale2016towards}. An extra class is added for the prediction, to model the unknown class case. OpenMax adapts meta-recognition concepts to the activation patterns in the penultimate layer of the network for unknown modelling. The Extreme Value Theory (EVT) is used  to estimate the probability of the input being an outlier. Several works have shown that in many cases easy strategies that look at the softmax probability or the logits can also effectively judge if the sample comes from an unknown class \cite{gavarini2022open}, e.g. exploiting the fact that the maximum output score tends to be smaller for inputs from unknown classes (out-of-set) \cite{vaze2022open, chow1970optimum}, or that the energy of the logit vector tends to be lower for out-of-set samples \cite{liu2020energy}.

	Other approaches explored reconstruction errors obtained via autoencoders for open set rejection  \cite{yoshihashi2019classification,oza2019c2ae,miller2021class}. In \cite{chen2020learning}, Yang et al. designed a suitable embedding space for open set recognition using convolutional prototype learning, that abandons softmax, and implements classification by finding the nearest prototype in the Euclidean norm in the feature space (GCPL). In \cite{yang2018robust}, a novel learning framework for OSR, called reciprocal point learning (RPL), is proposed. The method is extended in \cite{chen2021adversarial} (ARPL) via an adversarial mechanism that generates confusing training samples, to enhance the distinguishability of known and unknown classes. Recently, in \cite{huang2022class}, a method that combines autoencoders with, respectively, prototype learning (PCSSR) and reciprocal point learning (RCSSR) has been proposed.
	
	To the best of our knowledge, in the literature pertaining to synthetic manipulation detection, all the methods proposed so far are limited to the closed-set scenario. An exception is represented by \cite{girish2021towards}, where an algorithm to discover new GANs from a given unlabeled set and cluster them is proposed. More specifically, in the method in \cite{girish2021towards}, unseen classes are considered during network training as a unique -unlabeled - class (discovery set). The images belonging to this class are then clustered using the learned features, attributing them to new labels.

	\subsubsection{Vision Transformers for OSR}
	Transformers were originally proposed for natural language processing (NLP). Recently, they have been successfully exploited in computer vision tasks with great results \cite{han2022survey}. The extension of transformers to the image domain, namely, Vision Transformers (ViT) \cite{dosovitskiy2020image}, are self-attention architectures that process the image as a sequence of image patches, that are treated the same way as tokens (words) in the NLP case. Following the paradigm of the ViT architecture in \cite{dosovitskiy2020image}, a series of variants of the original structures have been proposed to further improve the performance on image tasks.
	
	Recent literature on machine learning has shown that ViT can achieve very good performance for out-of-distribution detection in image classification tasks \cite{fort2021exploring}, outperforming standard CNNs. Among the works exploiting ViT for open set recognition, we mention \cite{cai2022open,al2022open}. The method in \cite{al2022open}, in particular, combines ViT with energy-based rejection for open set scene classification in remote sensing imagery. Following the above literature, in this paper, we propose to exploit ViT for open-set classification of synthetic image manipulation.  To the best of our knowledge,  this is the first attempt in this direction.
	
	%%%%%% ----------------------------------------------------------
	\section{Proposed method}
	\label{sec:method}
	The general problem of open set classification of synthetic manipulation addressed in this paper is illustrated in Figure 1. A forensic classifier with rejection option classifies the type of synthetic manipulation, among those in a kwown set, at the same time being capable to reject unknown samples, namely, samples that were subject to a different manipulation or generation procedure with respect to those in the known set.
 
	As we said, we focus on the case of facial manipulations. Formally, given a synthesized face image $x \in \mathbb{R}^{H \times W \times 3}$ (height = H, weight = W), the system assigns to $x$ a label $y$ and a mask $M$ associated to the manipulation. Given the set with the $N$ known manipulations considered during training, the predicted label may take $N+1$ values, where the $N+1$-th value identifies the rejection class (unknown manipulation). If we let $\hat{y}$ denote the output (predicted label) of the $N$-class classification network, the final classification function $\phi(x)$ takes the following expression: $\phi(x) = \hat{y}$ if $x$ is accepted as an in-set sample, or $\phi(x) = N+1$ otherwise. In the following, we denote with $p \in \mathbb{R}^N$ is the probability vector (after softmax) of the network associated with the $N$ classes in the closed-set.
	%
	%\BTcomm{rephrase below. You already used $y$ for the classification label. Here you should use another letter if you need it. }\JUN{indicate with $s$}
	%{\em The localization mask is predicted that $s =1$ if $x$ is a manipulated image and $0$ otherwise, and the mask $M$ indicates the pixels where the image has been manipulated (pixels for which $M = 1$ indicate the manipulated areas).}
	%
	The localization mask $M$ is used to indicate the pixels where the image has been manipulated (pixels for which $M = 1$ indicate the manipulated areas). More in general, a localization mask may just highlight regions of interest (like an attention mask), without necessarily corresponding to a manipulation mask.

        \begin{figure}
            \centering
            \includegraphics[width= \columnwidth]{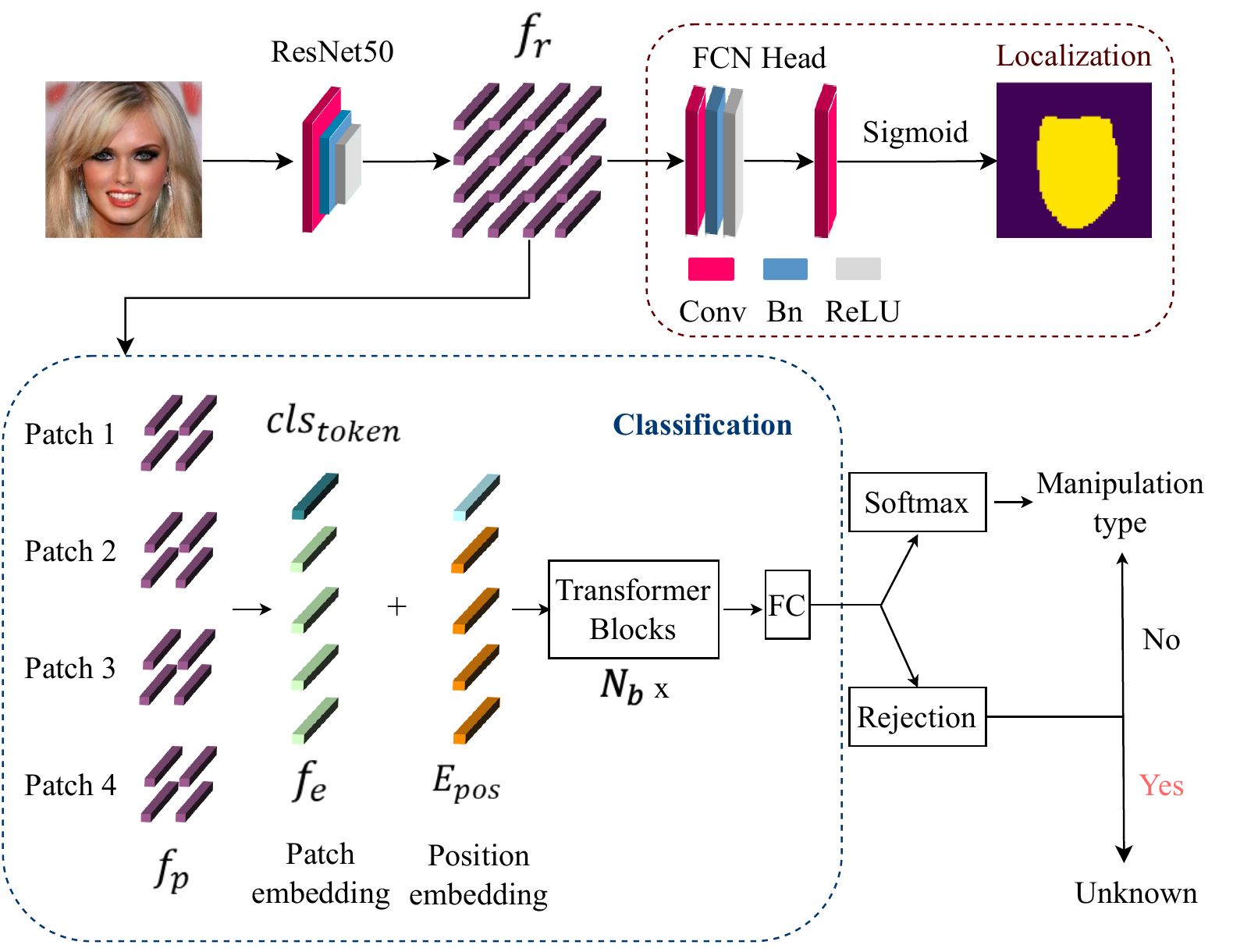}
            \caption{Overall architecture  of the proposed method.}
            \label{fig:framework}
        \end{figure}

	\subsection{Proposed ViT-based Hybrid Architecture}
	The general scheme of the proposed method is shown in Figure \ref{fig:framework}. The network is composed of two branches for classification and localization, respectively. A ResNet50 network is used as the backbone for feature extraction. Following \cite{gragnaniello2021gan,wang2022eyes}, a modification of the original ResNet architecture is considered, where we remove the sampling operation in the first convolutional layer of the network, setting the stride parameter to 1, with the  kernel size fixed to 3. The features are then input to a transformer-based module performing the $N$-class classification and to a fully convolutional network (FCN) head for the localization, as detailed below. Hence, in our scheme, the input sequence to the ViT is formed from feature maps of the CNN, as an alternative to raw image patches \cite{dosovitskiy2020image}.
	
	{\bf ViT-based classification module.}
	Let $f_r$ denote the vector of extracted features. We indicate with $(H_f, W_f)$ the size of the feature maps, and with $D_f$ the number of the channels/maps. Then, $f_r \in \mathbb{R}^{H_f \times W_f \times D_f}$. The following preprocessing is applied before feeding the ViT module. For a given patch size $P$, $f_r$ is first reshaped into a sequence of $P \times P \times D_f$ patches, in number $N_p = H_f W_f/P^2$. The special case $P=1$ corresponds to the case when the input sequence is obtained by simply flattening the spatial dimensions of the feature map and projecting to the transformer dimension. The input sequence obtained after this flattening layers is $f_p \in \mathbb{R}^{N_p \times \left ( P^{2}  D_f \right )}$.
	Following the general procedure with ViT, patch embedding is performed via mapping to $D_p$  dimensions via linear projection. $f_e = f_p \cdot E_p$ is the output, of shape of $N_p \times D_p$, obtained after the patch embedding operation, where $E_P$ denotes the embedding matrix, $E_p \in \mathbb{R}^{\left ( P^{2}  D_f \right )\times D_p}$. A placeholder data structure $cls_{token}$, used to store information that is extracted from other tokens in the sequence $f_e$, is prepended to the beginning of input sequence $f_e$ (randomly initialized). Position embeddings $E_{pos} \in \mathbb{R}^{(N_p + 1) \times D_p }$ are added to the patch embeddings to retain positional information, thus getting the sequence of vectors $ \left \{ cls_{token}, f_e \right \} + E_{pos}$, that is then fed to a standard transformation encoder, like those used in NLP. The transformer encoder is composed of $N_b$ identical transformer blocks, each one constructed of alternating layers of multi-headed self-attention (MHA) and multi-layer perception (MLP) blocks,  with a layernorm (LN) applied before every block, followed by residual connections after every block, see \cite{dosovitskiy2020image} for more details. Finally, a fully connected layer is attached to the transformer encoder and outputs the predicted probability vector $p$ for the $N$ enclosed classes. 

	{\bf FCN localization module.}
	The extracted features $f_r$ are also input to an FCN computing the estimated manipulation mask $M$. Such FCN consists of two convolutional layers, a batch normalization layer, a ReLU layer and finally a sigmoid layer to map the values in the $\left [ 0, 1 \right ]$ range. As we mentioned, the main reason for the introduction of the localization branch is to guide the classification and help force the network to focus on the most significant parts of the image, in the case of local manipulation, that has been shown to have a beneficial effect on the classification accuracy and generalization capabilities \cite{wang2022architecture}.

	The overall $N$-class classification architecture is trained end-to-end by minimizing a combination of the cross-entropy (CE) loss, associated with the classification task, and the mean squared error (MSE) of localization, respectively. Formally, $loss_{hyb} = \lambda_{cls} \cdot \text{CE}(y, p) + \lambda_{loc} \cdot \text{MSE}(G, M)$, where $G$ denotes the ground truth localization mask and $\lambda_{cls}$ and $\lambda_{loc}$ balance the trade-off between localization and classification tasks.
	
	The impact of each part of the proposed architecture, and in particular, the localization branch and the ViT module, is assessed in the experiments. For the ViT, we considered $N_b = 4$ transformer blocks. Different patch sizes $P$ of the ViT  were  considered in our experiments.

	\subsection{Out-of-Set Rejection}
	In order to detect samples whose manipulations do not belong to the knowm set, we considered three rejection strategies, two of which perform the rejection by analyzing the model output after or before the softmax activation layer, namely maximum softmax probability (MSP) \cite{chow1970optimum} and maximum logits score (MLS) \cite{vaze2022open}, respectively,  and OpenMax \cite{bendale2016towards}.
 
	When MSP and MLS approaches are adopted, lower scores associated with the predicted class reflect the uncertainty of the network prediction, providing evidence of unknown classes (out-of-set). Then, the final output of our open set classifier is obtained as follows
        \begin{equation}
    	\phi\left ( x \right ) = \left\{\begin{matrix}
    	\hat{y}, & \text{if $\arg \max(h) > th$} \\
    	N+1, & \text{otherwise} \\
    	\end{matrix}\right.
	\end{equation}
	where $h$ is the model output (the softmax probability in the MSP, the logit scores in the MLS), and $th$ is a predefined threshold. When the OpenMax is adopted, then the output of the closed-set classifier is accepted ($\phi( x  ) =  \hat{y}$) if  $p_o < th'$, where $p_o$ is the probability of the sample being an outlier, estimated by the method, and $th'$ is the decision threshold. Otherwise, it is rejected ($\phi( x  ) =N+1$).

	%%%%%% ----------------------------------------------------------
	\section{Experimental Setup}
	\label{sec:exper}
	
	\begin{table}
		\renewcommand\arraystretch{1.75}
		\caption{Summary of the 19 editing classes (18 + 'None'). \vspace{-0.4cm}}
		\begin{center}
			\resizebox{\linewidth}{!}{
				\begin{tabular}{c|c}
					\hline
					Editing tools & Edit types \\
					\hline
					PTI & T0: None (Reconstructed)\\
					\hline
					InterfaceGAN & \makecell{ \textbf{Expression} (T1-T2): Smile, Not smile,\\ \textbf{Aging} (T3,T4): Old, Young} \\
					\hline
					StyleCLIP&\makecell{\textbf{Expression} (T5, T6): Angry, Surprised \\ \textbf{Hairstyle} (T7-T12): Afro, Purple\_hair, \\ Curly\_hair, Mohawk, Bobcut, Bowlcut \\ \textbf{Identity change} (T13-T18): Taylor\_swift, Beyonce, \\ Hilary\_clinton, Trump, Zuckerberg, Depp}\\
					\hline
			\end{tabular}}
			\label{tab:dataset}
		\end{center}
	\end{table}

	\subsection{Datasets}
\begin{figure}
		\centering
		\includegraphics[width=\linewidth]{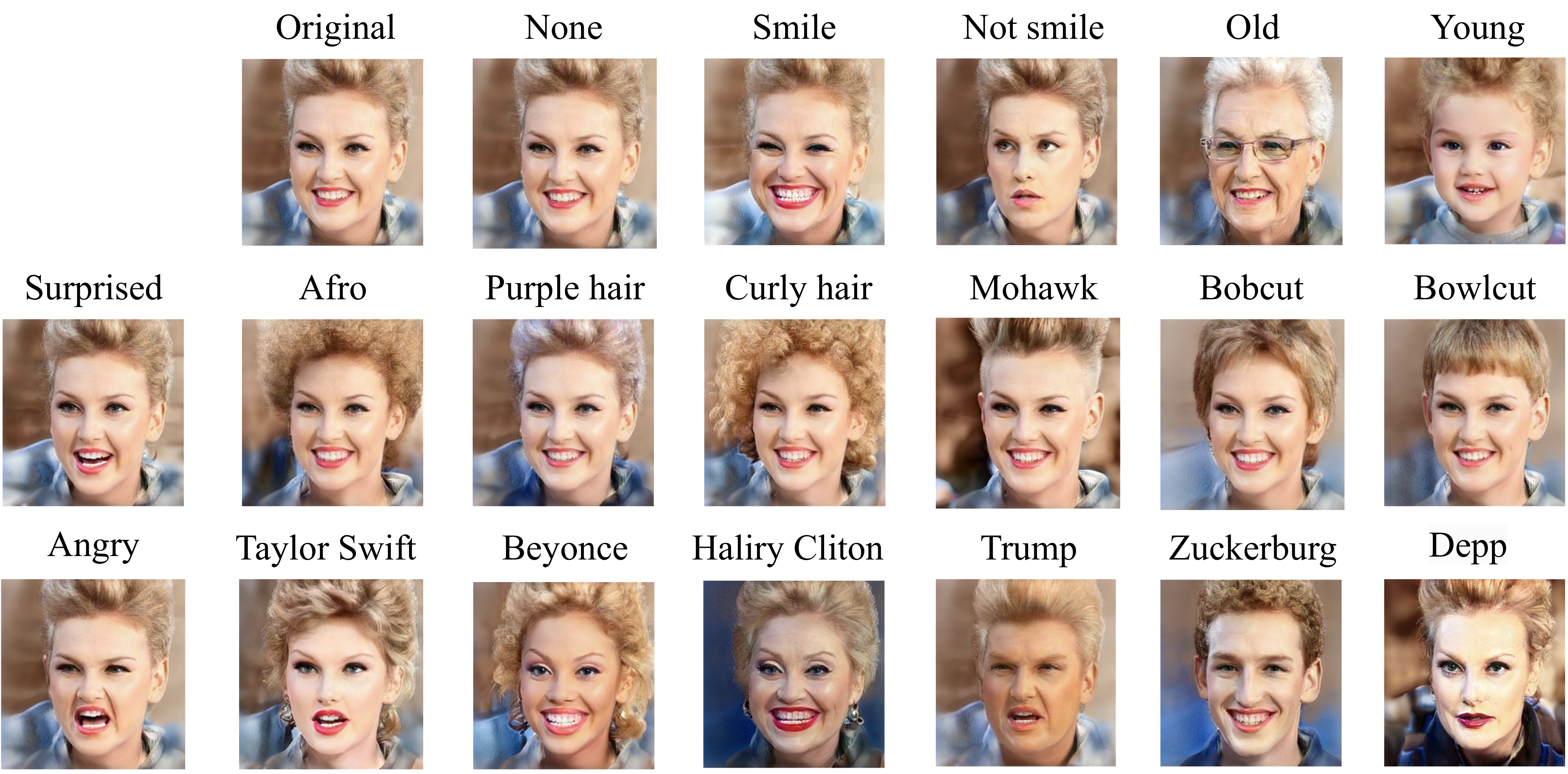}
		\caption{Examples of edited images by InterfaceGAN \cite{shen2020interpreting} (first row) and StyleCLIP \cite{patashnik2021styleclip} (second and third rows).}
		\label{fig:interfacegan}
	\end{figure}

	\textbf{GAN editing dataset.}
	To build the dataset, we first use the PTI inversion method to reconstruct the images and extract the latent code. Image attributes are manipulated by InterfaceGAN \cite{shen2020interpreting} and StyleCLIP \cite{patashnik2021styleclip}. We selected 5,992 images from CelebAHQ dataset and each image is edited with 18 edit types: 4 facial attributes are edited with InterfaceGAN, and 14 facial attributes with StyleCLIP. The 'None' type corresponds to the case of the image reconstructed with no editing (obtained via the PTI inversion method). An overview of our dataset is provided in Table \ref{tab:dataset}. We exploited a pre-trained face parsing model \cite{yu2018bisenet} to group the various edited attributes into four categories: expression, aging, hairstyle, identity change. We rely on these categories to construct the localization masks used for training. Figure \ref{fig:interfacegan} shows an example of manipulated face image for each edited attribute.
	
	\textbf{GAN attribution dataset.} To build the GAN attribution dataset used in our experiments we considered five GAN architectures: StyleGAN2 \cite{karras2020analyzing}, StyleGAN3 \cite{karras2021alias}, Taming Transformer \cite{esser2021taming}, Latent Diffusion \cite{rombach2022high} and LSGM \cite{vahdat2021score}. For each architecture, we considered 50k images. The models were trained on the FFHQ/FFHQU dataset. In all the cases, we used pre-trained models released by the authors. For LSGM, Taming transformers and Latent diffusion models, the resolutions of the images are 256 $\times$ 256, while for StyleGAN models the images are generated with both 256 $\times$ 256 and 1024$\times$1024 resolution.
	
	\subsection{Experimental setting}
	To train our model for GAN face editing classification, the dataset of real images is split as follows: $4400$ images are used to generate the editing used for training, $1592$ for those used for testing, for a total of $83600$ ($4400 \times 11$) images for training and $30248$ ($1592 \times 19$) for testing. Cross-validation is implemented during training by randomly splitting the training set in $4000 \times 11$ images used for training and $400 \times 11$ images used for validation, every 10 epochs. Training is performed via Adam optimizer with learning rate $lr = 10^{-5}$ and  batch size $bs =32$ for 100 epochs. The input size is set to 256$\times$256$\times$3. We ran comparison with the state-of-the-art methods in the field of OSR, i.e., GCPL \cite{chen2020learning}, RPL \cite{yang2018robust}, ARPL \cite{chen2021adversarial}, CAC \cite{miller2021class}, PCSSR and  RCSSR \cite{huang2022class}, mentioned in  Section \ref{sec.related}. All these methods are trained using the code released by the authors on our dataset with default setting and  input size  $224 \times 224$.
 
	As for GAN attribution, for each architecture, the images are split into 35000:5000:10000 for training, validation and testing, respectively. Training is carried out using the same optimizer, learning rate and batch size as above, for 50 epochs.
	Performance in the closed set is evaluated by measuring the classification accuracy, while the AUC of the ROC curve obtained by varying the thresholds  $th$ ($th'$ for OpenMax) is measured to evaluate the  rejection performance in  open set.
 
	% \subsubsection{Settings}
	\begin{figure}
		\centering
		\includegraphics[width=0.8\linewidth]{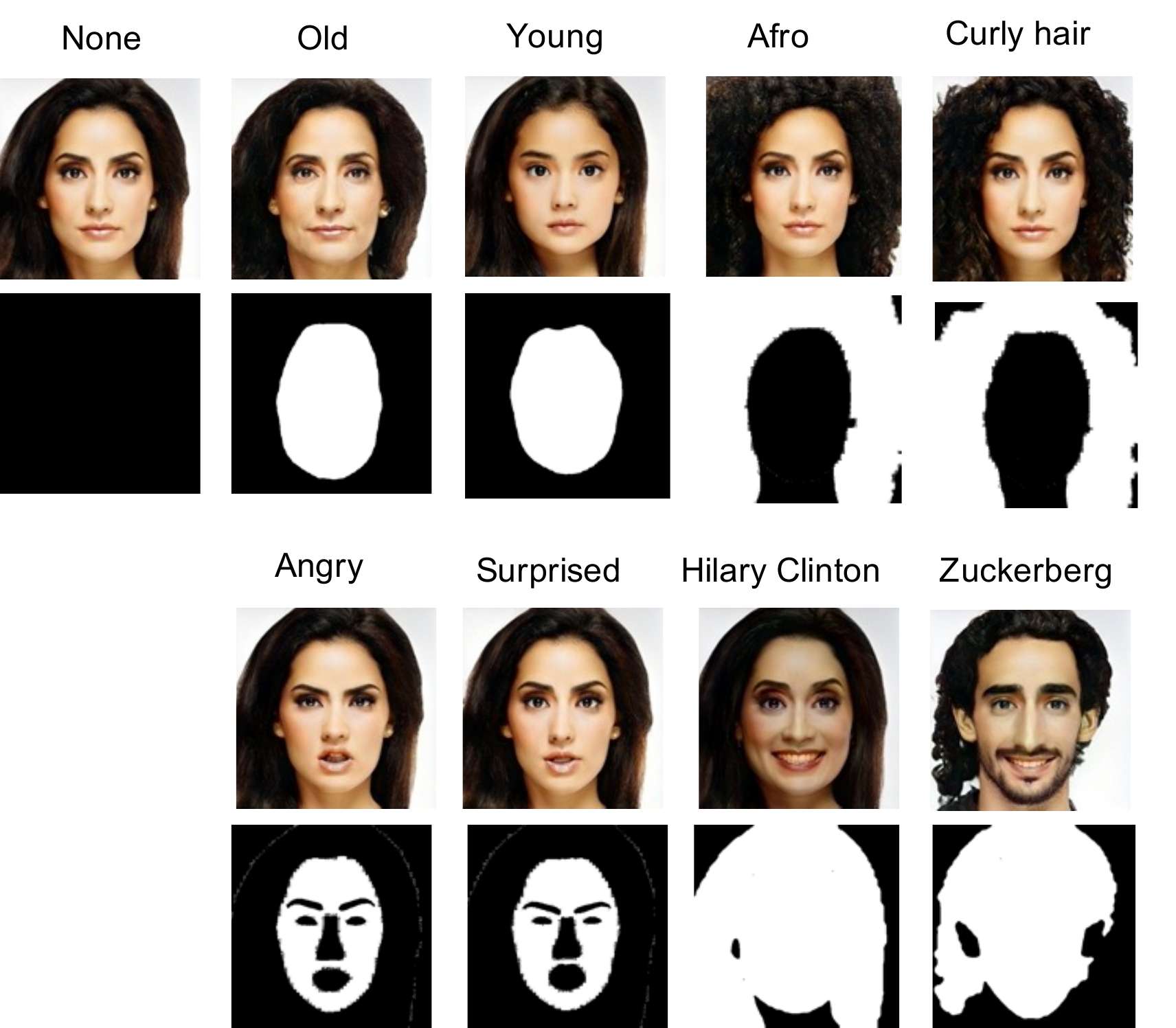}
		\caption{Examples of  images  and masks  obtained with different editing from each category. From left to right:  'None', aging (2),  hairstyle (2), expression (2), identity change (2).}
		\label{fig:masks}
	\end{figure}

	%\textbf{Mask Generation:}
	In the case of the classification of GAN face editing, the manipulation is performed locally and the localization branch is employed to guide the training. The ground truth localization masks highlighting the regions of interest in the images are used to train the model and are obtained as detailed in the following. We decided to use different masks for every category (expression, aging, hairstyle, identity change). We focus on the whole face area for aging editing while we consider the hair region for hairstyle editing. For identity editing, the focus area covers the whole face and hair since both of them are relevant in the characterization of identity. Finally, for expression editing, the profiles of the mouth, eyes, eyebrows and nose are enhanced in the masks by removing the corresponding segmented regions, being highly related to expressions. Some examples of masks are shown in Figure \ref{fig:masks}.
	
	%%%%%% ----------------------------------------------------------
	\section{Results}
	\label{sec:results}
	In this section, we report and discuss the results we got for the tasks of classification of GAN face editing and GAN attribution. Most of the experiments, in particular, the comparison with general state-of-the-art methods for OSR in machine learning, as well as an ablation study on the impact of the various elements of the proposed architecture and parameters, are reported for the former case. This is the case, in fact, where all the components of the proposed ViT-based hybrid network are considered, including the localization branch.
	
	\begin{table}
		\centering
		\renewcommand\arraystretch{1.3}
		\caption{Splitting of editing types considered in the experiments.}
		% \label{tab:freq}
  \small 
		\resizebox{\linewidth}{!}{
			\begin{tabular}{lcc}
				\toprule
				Groups & In-set & Out-of-set\\
				\midrule
				G0 & \makecell{T0, T2, T3, T5, T6, T7\\ T8, T9, T13, T14, T15} & \makecell{T1, T4, T10, T11 \\ T12, T16, T17, T18}\\
				\hline
				G1 & \makecell{T0, T1, T2, T5, T6, T13\\ T14, T15, T16, T17, T18} & \makecell{T4, T3, T7, T8 \\ T9, T10, T11, T12}\\
				\hline
				G2 & \makecell{T0, T1, T2, T5, T6, T7\\ T8, T9, T10, T11, T12} & \makecell{T4, T3, T13, T14\\ T15, T16, T17, T18}\\
				\hline
				G3 & \makecell{T0, T1, T2, T3, T4, T11\\ T12, T13, T14, T15, T18} & \makecell{T5, T6, T7, T8\\ T9, T10, T16, T17}\\
				\hline
				G4 & \makecell{T0, T1, T3, T4, T6, T10\\ T12, T15, T16, T17, T18} & \makecell{T2, T5, T7, T8\\ T9, T11, T13, T14}\\
				\bottomrule
		\end{tabular}}\label{tab:set}
	\end{table}

	\subsection{GAN face editing classification}
	Experiments were carried out considering 10 different configurations of in-set and out-of-set editing types, referred to as G0-G9.  In each case, 11 editing types are considered as in-set classes, while the remaining 8 are taken out-of-set.  Table \ref{tab:set} reports G0-G4 configurations, with the 'None' class always included as in-set. 
    Configurations G5-G9 are obtained from G0-G4 by switching the first in-set and out-of-set type, hence with the 'None' class in the out-of-set.

	Table \ref{tab:rejection} reports the closed-set accuracy and the open-set performance achieved with the 3 rejection strategies, for the various configurations. The average accuracy of the classification on the $N=11$ closed-set classes in the closed-set is 92.86. Regarding the open-set performance, the MLS is the strategy that gives the best results. In particular, with MLS we got AUC = 88.74 on average, in contrast to 81.19 and 81.86 for MSP and OpenMax respectively. Notably, the configurations for which the best closed-set performance is achieved correspond to those, that perform better in the open-set scenario. Therefore, in the following, results are reported for the MLS strategy, unless stated otherwise.
	
\begin{figure}
		\centering
		\includegraphics[width=\columnwidth]{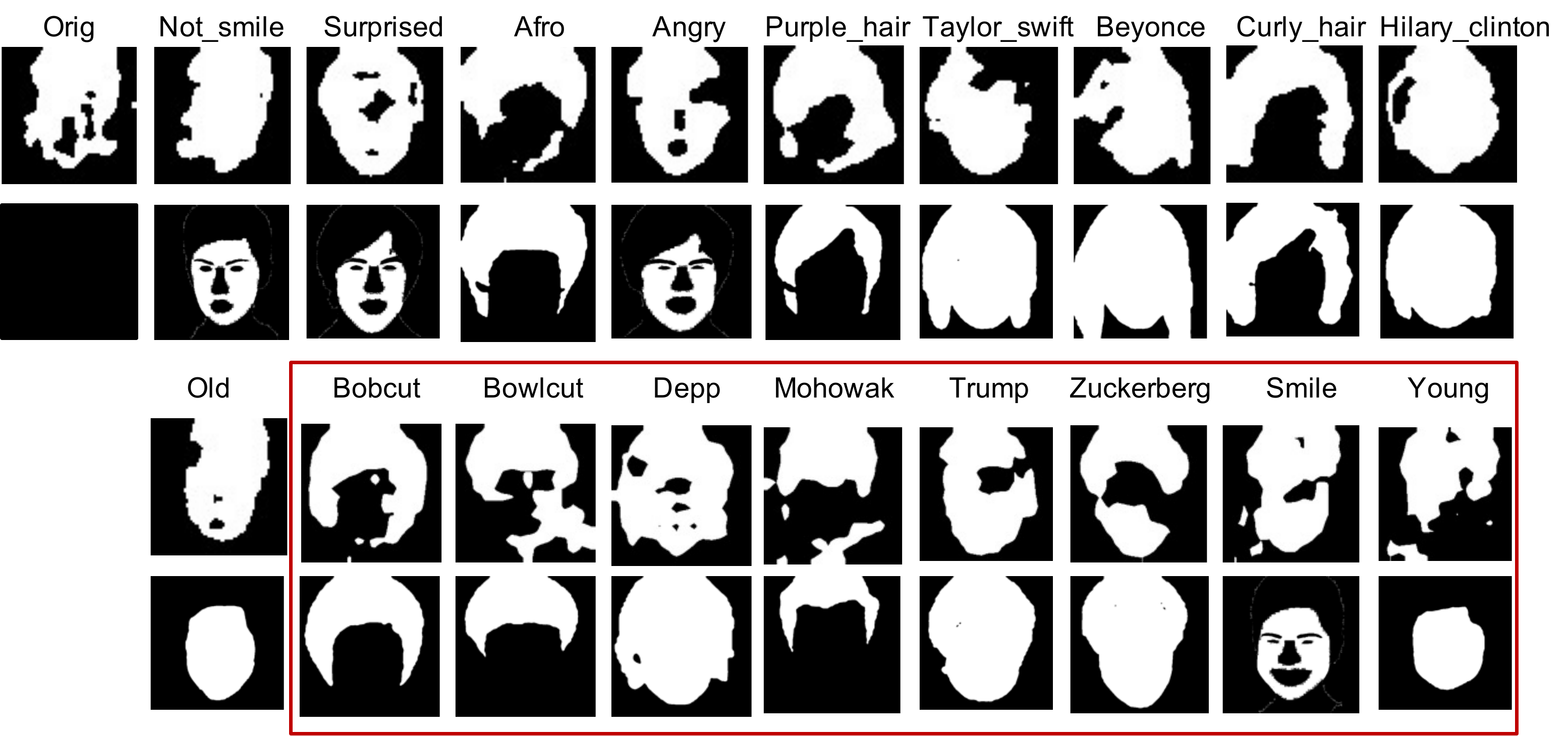}
		\caption{Example of localization masks for the 18 editing types. Predicted (top)  and ground truth (bottom) masks are visualized. The masks in the red box refer to the out-of-set editing types.}
		\label{fig:predmask}
	\end{figure}

	% In the results reported in the following, we consider the MLS  strategy to implement the rejection.
	In Figure \ref{fig:predmask}, we report an example of predicted masks in the various cases, for the G0 configuration, for visual assessment. Although the localization has been considered only to supervise the training, like an attention mechanism, and not for localization purposes, by looking at the figure, we can observe that in many cases the method is able to produce similar masks, namely masks with a similar white region (focus area), in both closed-set and open-set images, for editing types belonging to the same category (see Table \ref{tab:dataset}). This indicates that the network tends to look at areas of the image that are most relevant for the discrimination of the manipulation, and also for open-set inputs.

	The comparison of the proposed method with state-of-the-art algorithms is reported in Table \ref{tab:sota}, for the configurations G0, G3 and G4. We see that the proposed method achieves the best results in all the cases in both closed and open-set. In particular, our ViT-based hybrid algorithm gets an AUC of 85.34, 91.98 and 89.75 in G0, G3 and G4, respectively, getting an improvement with respect to the best-performing method from the state-of-the-art always larger than 4\% in both Accuracy and AUC. It is worth observing that all these methods have been proposed to address general problems of OSR in deep learning, and adopted for standard image classification tasks and object recognition, e.g. MNIST or CIFAR classification. Hence, they are not designed for forensic problems and in particular manipulation classification tasks, where the classification often relies on the analysis of subtle traces, and the goal in the open set scenario is being able to reveal unseen alterations of similar content or the presence of different fingerprints.

	\begin{table}
		\caption{Performance in closed-set and open-set, using different rejection strategies, for different configurations (G0-G9). The AUC is reported in the open-set.}
		\centering
		%\renewcommand\arraystretch{0.8}
		% \label{tab:freq}
		\resizebox{\linewidth}{!}{
			\begin{tabular}{llccccc}
				\toprule
				%  & MSP & OpenMax & MLS\\
				% \midrule
				% ours & 79.35 & 82.14 & 85.34\\
				Config & & G0 & G1 & G2 & G3 & G4 \\
				\midrule
				Closed-set & Accuracy & 88.99 & 94.68 & 87.03 & 94.34 & 95.25\\ 
				\hline
				\multirow{3}{*}{Open-set}& MSP & 79.35 & 79.63 & 71.49 & 84.54 & 83.97\\
				% \midrule
				& OpenMax & 81.83 & 81.89 & \textbf{81.39} & 74.86 & 81.34\\
				% \midrule
				& MLS & \textbf{85.34} & \textbf{91.36} & 78.34 & \textbf{91.98} & \textbf{89.75}\\
				\midrule
				\midrule
				Config & & G5 & G6 & G7 & G8 & G9 \\
				\hline
				{Closed-set} & Accuracy & 92.65 & 95.51 & 89.24 & 94.94 & 95.94\\ 
				\hline
				\multirow{3}{*}{Open-set}& MSP & 82.29 & 87.29 & 75.50 & 84.49 & 83.30\\
				& OpenMax & 78.62 & 86.20 & \textbf{83.72} & 85.00 & 83.73\\
				& MLS & \textbf{88.05} & \textbf{95.23} & 82.43 & \textbf{93.13} & \textbf{91.77}\\
				\bottomrule
			\end{tabular}
		}
		\label{tab:rejection}
	\end{table}
	
	\begin{table*}
		\centering
		\caption{Comparison with state-of-the-art method. Results are reported for the  G0, G3 and G4 configurations.}
      \vspace{-0.2cm}
	%	\Huge 
	%	\resizebox{\linewidth}{!}{
			\begin{tabular}{lcccccc}
				\toprule
				\multirow{2}{*}{Methods} & \multicolumn{2}{c}{G0} & \multicolumn{2}{c}{G3} & \multicolumn{2}{c}{G4}\\
				\cline{2-7}
				& Closed-set  & Open-set & Closed-set  & Open-set & Closed-set  & Open-set\\
				& (Accuracy) & (AUC) & (Accuracy) & (AUC) & (Accuracy) & (AUC)\\
				\midrule
%				CROSR \cite{yoshihashi2019classification} & 62.31 & 60.94 & & \\
				GCPL \cite{chen2020learning}& 73.72 & 73.25 & 40.93& 69.46 & 43.16 & 65.48\\
				RPL \cite{yang2018robust}& 74.43 & 76.21 & 70.19 & 81.46 & 65.76 & 71.18\\
				% ARPL \cite{chen2021adversarial}& 77.14 & 77.98 & 88.30 & 85.28 & 86.99 & 79.27\\
			%	ARPL+CS \cite{chen2021adversarial}
   ARPL \cite{chen2021adversarial} & 82.64 & 81.73 & 87.80 & 84.93 & 90.7 & 79.89\\
				CAC \cite{miller2021class}& 77.86 & 74.95 & 83.33 & 78.57 & 85.09 & 77.63\\
				PCSSR \cite{huang2022class} & 84.10 & 74.49 & 90.79 & 85.42 & 92.25 & 83.63\\
				RCSSR \cite{huang2022class} & 83.70 & 72.95 & 90.60 & 86.87 & 91.67 & 85.32\\
				 %MPF \cite{xia2021adversarial} & 79.91 & 78.12 & 90.03 & 87.86 & 91.34 & 81.39\\
				% AMPF \cite{xia2021adversarial} & 87.89 & 82.73 & 92.96 & 88.96 & 93.26 & 82.84\\
			%	AMPF++ \cite{xia2021adversarial} & \textbf{89.74} & 83.29 & 94.28 & 89.40 & 94.74 & 83.94\\
				% ours+MSP & \textbf{88.99} & \textbf{85.34}\\
				% ours+OpenMax & \textbf{88.99} & \textbf{85.34}\\
				Ours  & \textbf{88.99} & \textbf{85.34} & \textbf{94.34} & \textbf{91.98} & \textbf{95.25} & \textbf{89.75}\\
				\bottomrule
		\end{tabular}
	%}
		\label{tab:sota}
	\end{table*}

	\subsubsection{Ablation Study}
	
	We conducted an ablation study to investigate the effects of the patch size $P$ used in the ViT module and to validate the effectiveness of each component of the proposed architecture. 
	
\begin{figure}
		\centering
		\includegraphics[width=1.\columnwidth]{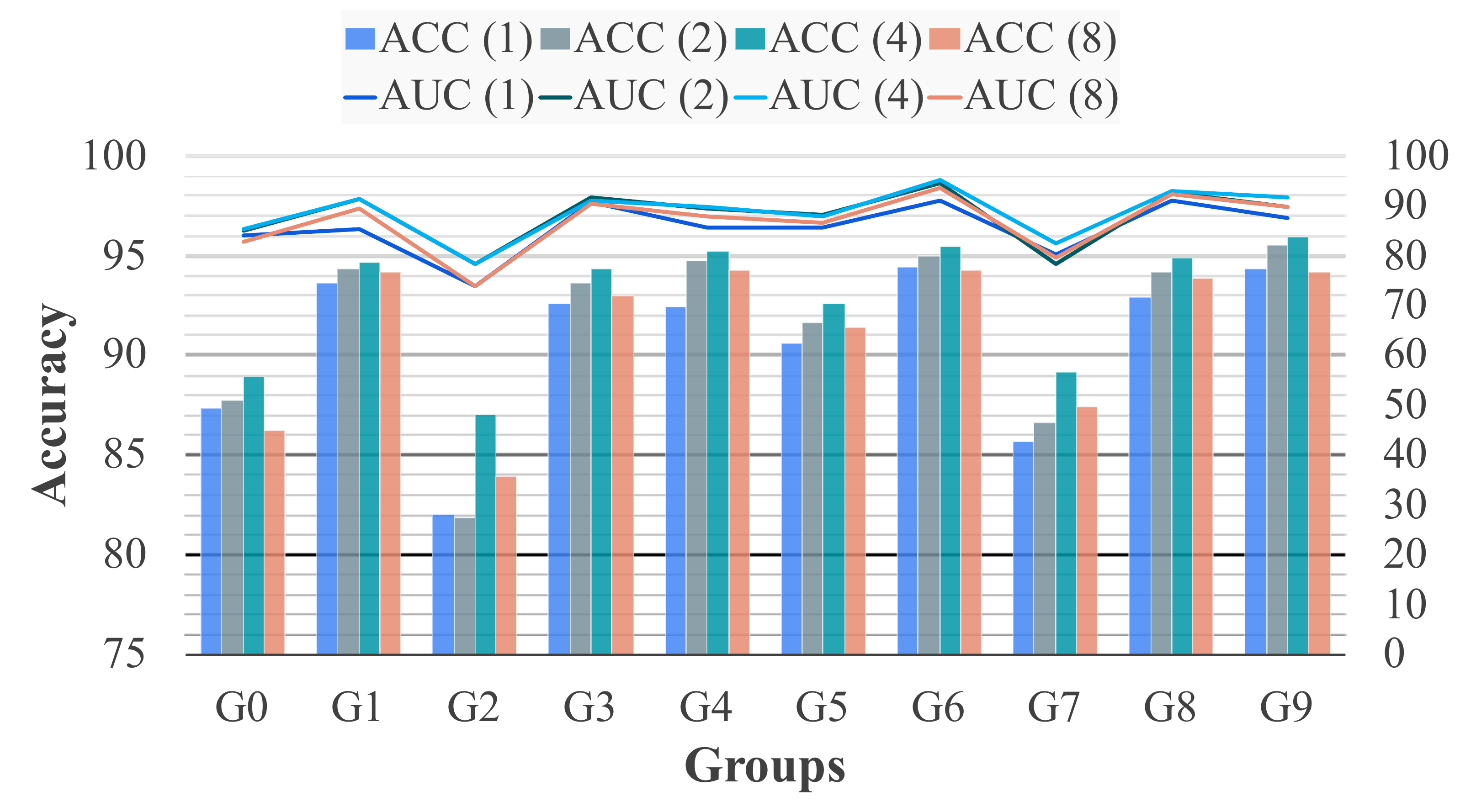}
		\caption{Ablation study on the impact of patch size $P$ of ViT under the various configurations. Vertical bars show closed-set Accuracy, while the line plots show the AUC for open-set.}
		\label{fig:ablapatch}
	\end{figure}

	\textbf{Impact of different patch sizes.} Figure \ref{fig:ablapatch} shows the results using different patch sizes $P$ in the ViT module, namely $P =$ 1, 2,  4 and 8 (the legends reports the  $P$ setting among brackets). 
 %We see that using larger patches, namely $P = 4$, is beneficial for both closed-set and open-set performance.
 We see that increasing the patch size, up to $P = 4$, is beneficial for both closed-set and open-set performance.
% In our experiments, we have also observed that
However, when the patch size increases further, namely, above 4,  results do not improve, and actually  a performance drop is observed  (of around 1.6\% in  Accuracy and 2\% in AUC on the average). Then, from our experiments, with $P = $4 the ViT achieves the best tradeoff between the exploitation of the spatial and of the feature maps correlation.
	
\begin{figure*}
		\centering
		\includegraphics[width=1\columnwidth]{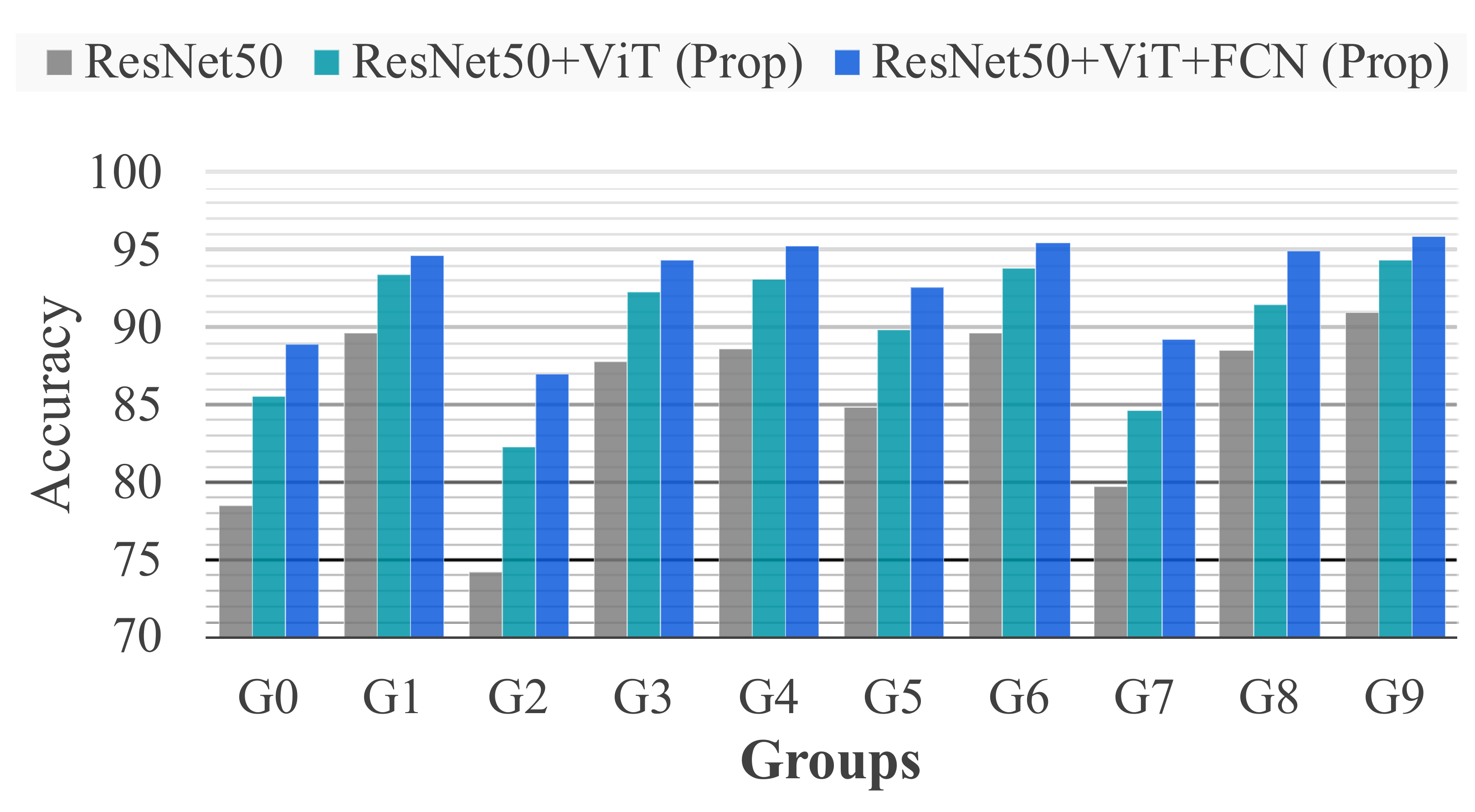}
		\includegraphics[width=\columnwidth]{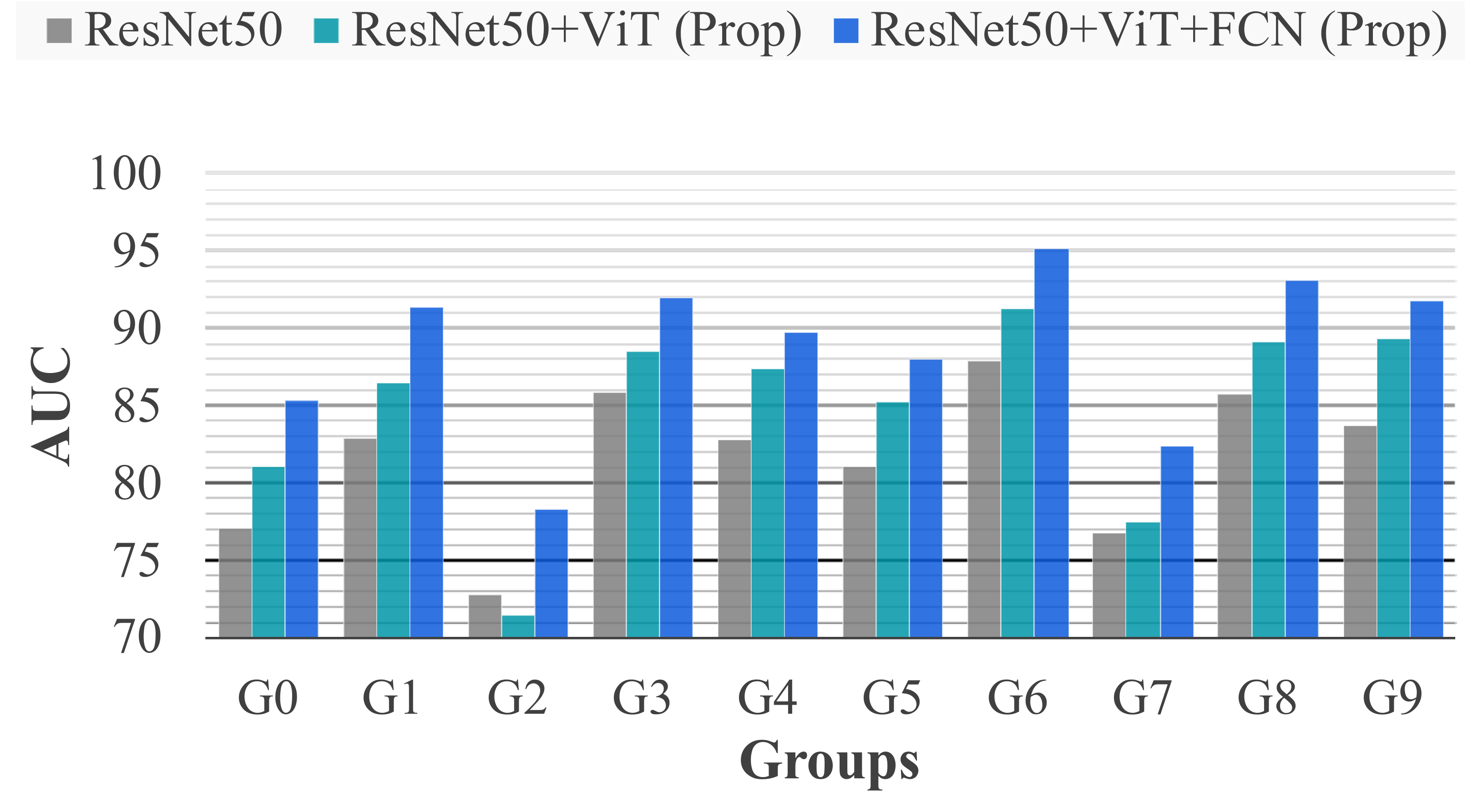}
  \vspace{-0.3cm}
		\caption{ Performance in closed-set (left) and open-set (right) for different configurations (G0-G9).}
		\label{fig:ablation}
	\end{figure*}

	\textbf{Impact of different architectures} 
	Figure \ref{fig:ablation} compares the results achieved by the proposed architecture including the ViT module for the classification and the localization branch (FCN), with those achieved by the same method by removing the FCN, and those of the baseline ResNet50, where the standard ResNet50 is used for the multi-class classification. In this case, the rejection is performed in a similar way, by analyzing the output layer of the last FC of ResNet50, before the softmax (MLS). A significant performance gain is obtained with the proposed method in all the configurations. In particular, combining the use of ViT for processing the feature maps with the hybrid approach we got a gain in performance of up to 10\% in Accuracy and 9\% in AUC.
	
	\subsection{GAN attribution}

	In this section, we report the results we got for open-set GAN attribution. By focusing on fully synthetic images, we do not include the localization branch in the proposed architecture, but only the ViT module. Experiments are carried out considering 4 different splittings of the 5 architectures. The in-set and out-of-set architectures for each configuration are detailed in the following: S1) in-set: LSGM, StyleGAN2 and Taming transformer; out-of-set: StyleGAN3 and Latent diffusion; S2) in-set: StyleGAN2, StyleGAN3 and Latent diffusion; out-of-set: LSGM and Taming transformer; S3) in-set: LSGM, StyleGAN2 and StyleGAN3; out-of-set: Taming transformer and Latent diffusion; S4) in-set: LSGM, StyleGAN2 and Latent diffusion; out-of-set: StyleGAN3 and Taming transformer. 

 	\begin{table}
		\caption{Results on GAN attribution task.}
  \vspace{-0.2cm}
		% \label{tab:freq}
		\centering
		\Huge
		\renewcommand\arraystretch{1.25}
		\resizebox{\linewidth}{!}{
			\begin{tabular}{cccccc}
				\toprule
				\multirow{2}{*}{Config} & \multirow{2}{*}{Method} & \multirow{2}{*}{\makecell{Closed-set\\ (Accuracy)}} &\multicolumn{3}{c}{Open-set (AUC)}\\
				\cline{4-6}
				& & & MSP & OpenMax & MLS\\
				\midrule
				\multirow{2}{*}{S1} 
				& ResNet50 & 97.76 & 77.23 & 64.60 & 76.32\\
				& ResNet50+ViT (prop)  & \textbf{99.86} & \textbf{92.73} & \textbf{92.70} & \textbf{92.72}\\
				% & Proposed hybrid network & \textbf{99.94} & 93.31 & 63.97 & 89.41\\
				\hline
				\multirow{ 2}{*}{S2} 
				& ResNet50 & 78.26 & 39.90 & 33.40 & 43.40 \\
				& ResNet50+ViT (prop) & \textbf{82.38} & \textbf{72.49} & \textbf{65.02} & \textbf{70.39}\\
				%& Proposed hybrid network & 80.82 & 65.11 & 53.64 & 68.05\\
				\hline
				\multirow{2}{*}{S3} 
				& ResNet50 & 92.80 & 78.78 & 57.83 & 69.82 \\
				& ResNet50+ViT (prop) & \textbf{98.56} & \textbf{82.74} & \textbf{78.13} & \textbf{83.31}\\
				%& Proposed hybrid network & 95.87& \textbf{86.46}& 76.68 & 82.80\\
				\hline
				\multirow{2}{*}{S4} 
				& ResNet50 & 81.82 & 67.43 & 61.66 & 69.98 \\
				& ResNet50+ViT (prop) & \textbf{94.61} & \textbf{90.31} & \textbf{93.56} & \textbf{93.60}\\
				%& Proposed hybrid network & \textbf{95.45} & 83.28& \textbf{93.93} & 88.90\\
				\bottomrule
		\end{tabular}}\label{tab:ganattri}
	\end{table}
	% Preliminary results, totally five GAN architectures are tested. Trained on three and test on five, with two as open set rejection testing.
 
	Table \ref{tab:ganattri} shows the closed-set and open-set performance achieved by the proposed architecture (\textit{ResNet50+ViT}) in all the configurations. The results of the baseline are also reported (\textit{ResNet50}). We see that the advantage we got with respect to the baseline is even bigger in this case than that in the previous case of face editing classification. In particular, when the rejection strategies are mounted on top of the baseline architecture, that is, by considering  the features extracted by a standard ResNet50 classifier for the  analysis, the rejection performance is very poor, with an AUC lower than 70\% in most cases. Our method instead can achieve a much higher AUC, that goes above 90 \% for S1 and S4. Under the S1 and S2 configurations, the results are worse. We observe that these configurations include both StyleGAN2 and 3 in the training set, hence resulting in a lower  diversity of the in-set dataset, that might be the reason for the worse generalization capability to open-set scenarios. Finally, we observe that, as before, the MLS is the strategy that gives the best performance on average, even if in this case the 3 rejection strategies work very similarly. These results confirm that the features extracted with our architecture are representative and allow a good characterization of the various architectures, yielding  good discrimination  also in the open-set scenario.

	%%%%%% ----------------------------------------------------------
	\section{Conclusion}
	\label{sec:conclusion}
% \MB{In general I do not like writing the Conclusions as a summary of what e have said already, a kind of abstract written in past tense. Isn't it possible to give some more insightful perspectives on the approach we used and the assumptions we made to develop our method? Future works are OK.}\JUN{See in italic, a minor change} \MB{It doesn't change much indeed. Given that the scheme we are proposing adopts several strategies to improve performance,. like ViT, localization, a number of rejection options, why don't you briefly discuss which of the above is more important to improve the performance from different perspectives, i.e., classification accuracy and rejection accuracy?}\JUN{I have tried to rewrite the conclusion part.}
%
        % \textit{We extend and explore new use cases of the ViT module in open-set classification for GAN-based image manipulation. Deep associations between feature patches explored by Vit significantly improve closed-set accuracy and open-set rejection ability, which is further boosted by manipulation localization, forcing the feature extraction layers to focus on the most likely manipulated parts of the image. Furthermore, rejection is performed using the logit output of multiple classifiers, avoiding the normalization operation of softmax, with great success on the open set.} Experiments demonstrate the effectiveness of the proposed method, also compared to other state-of-the-art methods for open-set classification, for the task of classification of GAN face editing and GAN attribution.   
	We have presented a method to address the problem of open-set classification of synthetic manipulations. A multi-class classifier with rejection option is implemented, that classifies the manipulation, at the same time being capable to reject an unknown (unseen) manipulation. {To address this task, we resort to a ViT-based hybrid architecture that explores global attention from patches while being guided by manipulation localization.}
    % combines the use of ViT with an approach based on simultaneous classification and localization. 
    Rejection is performed via several approaches, that rely on the analysis of the output logits and scores, and on outlier probability estimation.
	Experiments demonstrate the effectiveness of the proposed method,   also compared to other state-of-the-art methods for open-set classification,  for the task of classification of GAN face editing and GAN attribution. 
	
	Future works  will focus on the application of the proposed architecture to different synthetic manipulation classification tasks, considering different image contents, beyond faces. The robustness of the proposed method against post-processing and attacks is also worth investigating. Moreover, the promising results achieved for GAN attribution encourage us to explore further the use of the proposed architecture for this task.

	%problem in a multi-class classification task of synthetic images with edited facial attributes, which involves unknown classes that appear in testing while not considered during training. The goal of this task is to classify a facial attribute edited image with a correct manipulation type for the known classes and to reject prediction for unknown classes from open set. 
	
	%To tackle the challenging task, we presented a Vit-based hybrid classification and localization baseline model for the facial attribute editing task. The Vit module is mainly used to extract features among patches of feature maps and is guided by the localization task. Afterwards, the maximum logit score of the Vit module is considered for unknown class rejection. The numerical results on facial manipulation show that the proposed architecture outperforms state-of-the-art methods in the field of open set recognition. Furthermore, the proposed method is applied to open set GAN attribution task where the power of the Vit module is confirmed for the open set problem. 

	%We hope that this work draws the attention of researchers to open-set forensics problems beyond simple detection/classification tasks. In future work, we would collect a large-scale dataset that includes diverse image domains of GAN edited/synthesized images, which will allow us to tackle more realistic tasks in challenging scenarios.

    \section*{Acknowledgement}
    This work has been partially supported by the China Scholarship Council (CSC), file No. 202008370186, by the PREMIER project under contract PRIN 2017 2017Z595XS-001, funded by the Italian Ministry of University and Research, and by the Defense Advanced Research Projects Agency (DARPA) and the Air Force Research Laboratory (AFRL) under agreement number FA8750-20-2-1004. 
    The U.S. Government is authorized to reproduce and distribute reprints for Governmental purposes notwithstanding any copyright notation thereon.
	
	\noindent
	
	%%%%%%%%%
	% \clearpage
	%%%%%%%%% REFERENCES
	{\small
		\bibliographystyle{ieee_fullname}
		\bibliography{egbib}
	}
	
\end{document}